\documentclass[sigconf]{acmart}

\AtBeginDocument{%
  }

\copyrightyear{2026}
\acmYear{2026}
\setcopyright{cc}
\setcctype{by}
\acmConference[ICCAD '26]{IEEE/ACM International Conference on Computer-Aided Design}{November 08--12, 2026}{San Jose, CA, USA}
\acmBooktitle{IEEE/ACM International Conference on Computer-Aided Design (ICCAD '26), November 08--12, 2026, San Jose, CA, USA}
\acmDOI{10.1145/3831252.3834264}
\acmISBN{979-8-4007-2873-0/2026/11}

\usepackage{xcolor}
\usepackage{multirow}    
\usepackage{colortbl}
\usepackage{pifont}
\newcommand{\up}[1]{{\scriptsize\textcolor{green!70!black}{\,$\uparrow$#1}}}
\definecolor{bestgreen}{HTML}{C6EFCE}
\definecolor{failred}{HTML}{FFC7CE}

\begin{document}

\title[COEVO]{COEVO: Co-Evolutionary Framework for Joint Functional Correctness and PPA Optimization in LLM-Based RTL Code Generation}

\settopmatter{authorsperrow=4}

\author{Heng Ping}
\affiliation{%
  \institution{University of Southern California}
  \city{Los Angeles}
  \country{USA}}
\email{hping@usc.edu}

\author{Peiyu Zhang}
\affiliation{%
  \institution{University of Southern California}
  \city{Los Angeles}
  \country{USA}}
\email{pzhang65@usc.edu}

\author{Shixuan Li}
\affiliation{%
  \institution{University of Southern California}
  \city{Los Angeles}
  \country{USA}}
\email{sli97750@usc.edu}

\author{Wei Yang}
\affiliation{%
  \institution{University of Southern California}
  \city{Los Angeles}
  \country{USA}}
\email{wyang930@usc.edu}

\author{Anzhe Cheng}
\affiliation{%
  \institution{University of Southern California}
  \city{Los Angeles}
  \country{USA}}
\email{anzheche@usc.edu}

\author{Shukai Duan}
\affiliation{%
  \institution{University of Southern California}
  \city{Los Angeles}
  \country{USA}}
\email{shukaidu@usc.edu}

\author{Xiaole Zhang}
\affiliation{%
  \institution{University of Southern California}
  \city{Los Angeles}
  \country{USA}}
\email{xiaolezh@usc.edu}

\author{Paul Bogdan}
\affiliation{%
  \institution{University of Southern California}
  \city{Los Angeles}
  \country{USA}}
\email{pbogdan@usc.edu}

\renewcommand{\shortauthors}{Ping et al.}

\begin{abstract}
LLM-based RTL code generation methods increasingly target both functional correctness and PPA quality, yet existing approaches decouple the two objectives, optimizing PPA only after full correctness is achieved. Sequential multi-agent pipelines, evolutionary search with binary correctness gates, and hierarchical reward dependencies all discard partially correct but architecturally promising candidates. Moreover, existing methods reduce the multi-objective PPA space to a single scalar fitness that obscures the trade-offs among area, delay, and power. To address these limitations, we propose COEVO, a co-evolutionary framework that unifies correctness and PPA optimization within a single evolutionary loop. COEVO formulates correctness as a continuous co-optimization dimension alongside area, delay, and power, enabled by an enhanced testbench with fine-grained scoring. The framework allows PPA-promising but partially correct candidates to guide the search toward jointly optimal solutions, and further employs four-dimensional Pareto-based non-dominated sorting to preserve the PPA trade-off structure without manual weight tuning. On VerilogEval 2.0 and RTLLM 2.0, COEVO surpasses all agentic baselines across four LLM backbones. With GPT-5.4-mini, COEVO achieves 97.5\% and 94.5\% Pass@1 and attains the best PPA on 43 out of 49 synthesizable RTLLM 2.0 designs. Our code is available at \url{https://github.com/hping666/COEVO}.

\end{abstract}

\ccsdesc[300]{Computing methodologies~Artificial intelligence}
\ccsdesc[500]{Hardware~Hardware description languages and compilation}
\ccsdesc[500]{Hardware~High-level and register-transfer level synthesis}

\keywords{RTL code generation, large language models, co-evolutionary optimization, PPA optimization}

\maketitle


\section{Introduction}
\label{sec:intro}

The growing complexity of modern integrated circuits has made manual Register-Transfer Level (RTL) design a costly bottleneck in hardware development \cite{Liu2023ChipNemo, Chang2023ChipGPT}. Large language models (LLMs) \cite{yang2025toward} offer a promising direction for automating RTL code generation from natural language specifications \cite{Yang2025LLMVerilog, Xu2025LLMsEDA, Ping2025HDLCoRe}, with standardized benchmarks such as VerilogEval \cite{Liu2023VerilogEval, Pinckney2025RevisitingVerilogEval} and RTLLM \cite{Liu2024OpenLLMRTL, Lu2024RTLLM} accelerating progress. Extensive research has advanced functional correctness through domain-specific fine-tuning \cite{Liu2024RTLCoder, Pei2024BetterV, Zhao2025CodeV, Cui2024OriGen}, reinforcement learning with tool feedback \cite{Teng2025VERIRL, Wang2025VeriReason}, and multi-agent frameworks \cite{Liu2026VeriSure, Zhao2025MAGE, Ping2025VeriMoA, Deng2026ACERTL}. However, functional correctness alone is insufficient for practical deployment, as LLM-generated RTL frequently exhibits suboptimal PPA compared to engineer-written implementations \cite{Chen2025ChipSeekR1, Gubbi2025PromptingPower, Ping2026POET, Tasnia2025VeriOpt}. Jointly optimizing functional correctness and PPA quality during spec-to-RTL generation has therefore become an important research direction.

A growing body of work has begun to address PPA quality in LLM-based RTL design. RTL-to-RTL methods such as POET \cite{Ping2026POET} improve PPA on existing correct designs, but do not address the spec-to-RTL setting where no correct implementation is available. Within the spec-to-RTL setting, multi-agent frameworks such as VeriOpt \cite{Tasnia2025VeriOpt} assign dedicated agents for correctness and PPA. However, these sequential pipelines optimize PPA only after correctness is confirmed: conservative modifications yield limited PPA gains, while aggressive restructuring risks functional regression. Evolutionary approaches such as REvolution \cite{Min2026REvolution} broaden design space exploration through population-based search, but impose a binary correctness gate and collapse multi-objective PPA into a weighted-sum scalar. ChipSeek-R1 \cite{Chen2025ChipSeekR1} incorporates PPA feedback into reinforcement learning training but still suppresses PPA signals when correctness is not achieved. Across all paradigms, two fundamental limitations persist. First, correctness is decoupled from PPA optimization, so partially correct designs with superior architectural characteristics are systematically discarded rather than serving as stepping stones toward jointly optimal solutions. Second, multi-objective PPA is collapsed into a single scalar fitness. Scalarization obscures trade-offs among area, delay, and power \cite{Qi2024EvoT} and requires manual weight tuning without guaranteeing Pareto optimality.

To address both limitations, we propose COEVO, a co-evolutionary framework that unifies correctness and PPA optimization within a single evolutionary loop. To overcome the decoupled optimization problem, COEVO formulates correctness as a continuous co-optimization dimension alongside area, delay, and power. This continuous formulation is enabled by an enhanced testbench with fine-grained scoring and diagnostic feedback. An adaptive correctness gate and cross-objective operators allow PPA-promising but partially correct candidates to guide the search while converging to full correctness. To overcome single-objective scalarization, COEVO employs 4D (four-dimensional) Pareto-based non-dominated sorting \cite{Deb2002NSGAII} with configurable intra-level sorting that preserves trade-off structure without manual weight tuning.

Our contributions are summarized as follows:
\begin{itemize}
    \item We propose COEVO, a co-evolutionary framework for LLM-based RTL code generation that jointly optimizes functional correctness and PPA, where an adaptive correctness gate and cross-objective operators enable partially correct but PPA-promising candidates to serve as stepping stones.
    \item We introduce 4D Pareto-based non-dominated sorting with configurable intra-level sorting to the spec-to-RTL setting, preserving PPA trade-off structure and enabling flexible preference expression without manual weight tuning.
    \item Comprehensive experiments on VerilogEval 2.0 and RTLLM 2.0 demonstrate that COEVO achieves state-of-the-art functional correctness while producing designs with superior PPA metrics.
\end{itemize}

\section{Related Work}
\label{sec:related_work}

\subsection{LLM-Based RTL Code Generation}

Automating RTL code generation from natural language specifications has attracted significant research attention. Domain-specific fine-tuning methods include RTLCoder \cite{Liu2024RTLCoder}, BetterV \cite{Pei2024BetterV}, CodeV \cite{Zhao2025CodeV}, OriGen \cite{Cui2024OriGen}, and ScaleRTL \cite{Deng2025ScaleRTL}. Reinforcement learning methods such as VeriRL \cite{Teng2025VERIRL} and VeriReason \cite{Wang2025VeriReason} incorporate compiler and simulator feedback into training. In parallel, agentic frameworks leverage general-purpose LLMs without modifying model parameters. MAGE \cite{Zhao2025MAGE} and VerilogCoder \cite{Ho2025VerilogCoder} decompose generation into specialized roles, VeriMoA \cite{Ping2025VeriMoA} performs quality-guided mixture-of-agents generation, and VeriSure \cite{Liu2026VeriSure} integrates formal verification with temporal tracing. ACE-RTL \cite{Deng2026ACERTL} combines an RTL-specialized LLM with a frontier reasoning model. These approaches have advanced functional correctness but do not systematically improve PPA quality.

\subsection{PPA-Aware RTL Design with LLMs}

A growing body of work seeks to incorporate PPA optimization into LLM-based design flows. In the RTL-to-RTL setting, RTLRewriter \cite{Yao2024RTLRewriter} employs cost-aware MCTS for rewriting, SymRTLO \cite{Wang2025SymRTLO} combines LLM-driven dispatching with symbolic reasoning, and POET \cite{Ping2026POET} introduces power-oriented evolutionary optimization. These methods require a pre-existing correct implementation and do not address spec-to-RTL generation.
Within the spec-to-RTL setting, in-context learning methods such as Prompt for Power \cite{Gubbi2025PromptingPower} and LLM-VeriPPA \cite{Thorat2025LLMVeriPPA} inject PPA knowledge into prompts or leverage synthesis reports. Multi-agent frameworks such as VeriOpt \cite{Tasnia2025VeriOpt} and VeriAgent \cite{Wang2026VeriAgent} assign dedicated agents for correctness and PPA optimization. Evolutionary methods including REvolution \cite{Min2026REvolution}, EvolVE \cite{Hsin2026EvolVE}, and VFlow \cite{Wei2026VFlow} maintain populations and iteratively evolve designs. ChipSeek-R1 \cite{Chen2025ChipSeekR1} incorporates hierarchical PPA reward into reinforcement learning. Despite covering diverse paradigms, these methods share the two limitations discussed in Section~\ref{sec:intro}: they decouple correctness from PPA optimization and reduce multi-objective PPA to a single scalar fitness.

\subsection{Evolutionary LLM Frameworks}

Combining LLMs with evolutionary computation has emerged as a promising paradigm. FunSearch \cite{RomeraParedes2024FunSearch} evolves functions within program skeletons, EoH \cite{Liu2024EoH} co-evolves code and natural language heuristics, and EoT \cite{Qi2024EvoT} formulates LLM reasoning as multi-objective optimization using NSGA-II \cite{Deb2002NSGAII}. In the RTL domain, REvolution \cite{Min2026REvolution} couples evolutionary search with adaptive UCB-Softmax operator selection, and EvolVE \cite{Hsin2026EvolVE} guides the search with MCTS and structured testbench generation. VFlow \cite{Wei2026VFlow} instead searches over LLM-invoking workflow DAGs. A critical design choice is the fitness function for multi-objective PPA: REvolution uses a weighted sum with manual weights, EvolVE uses the area-delay product, VFlow decomposes into per-objective populations, and POET \cite{Ping2026POET} adopts non-dominated sorting with power-first ranking but operates in RTL-to-RTL. COEVO extends non-dominated sorting to spec-to-RTL by introducing correctness as a continuous fourth Pareto dimension with configurable intra-level sorting.

\begin{figure*}[t]
\centering
\includegraphics[width=\textwidth]{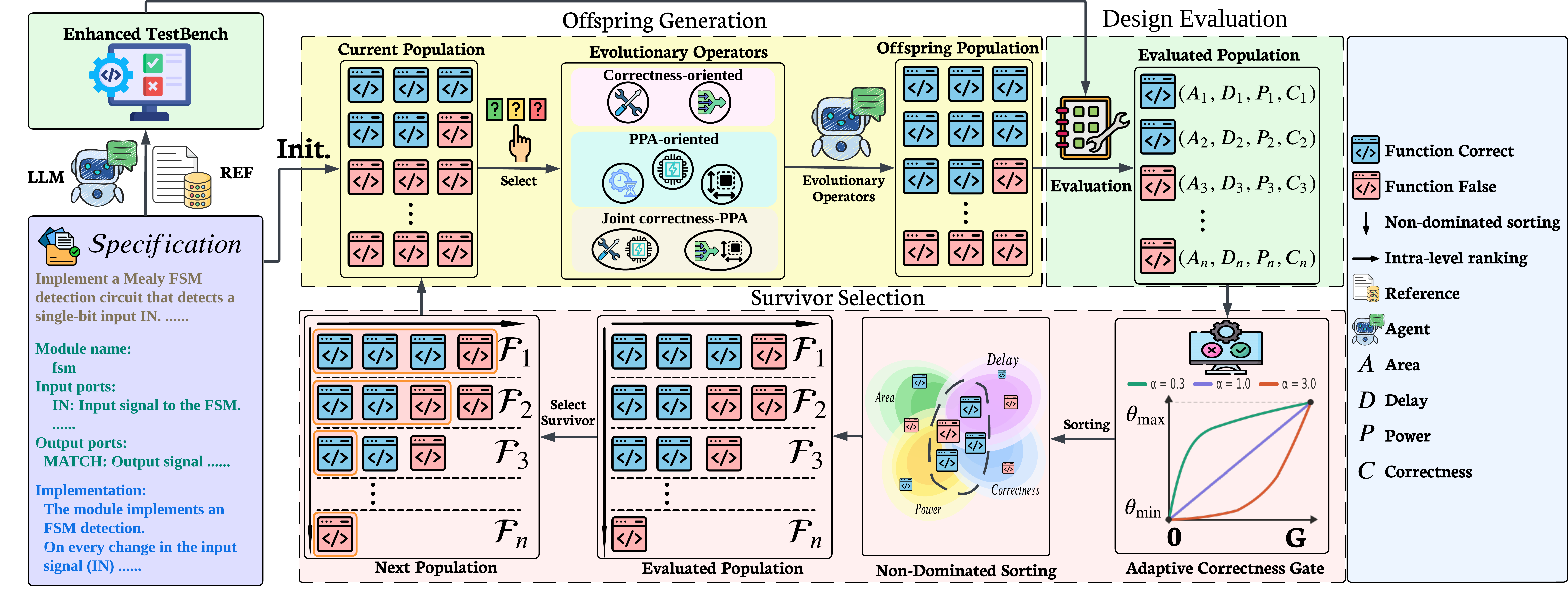}
\caption{Overall architecture of COEVO. The framework iteratively refines a population of design candidates through three stages: offspring generation via LLM-driven evolutionary operators, design evaluation for correctness and PPA, and survivor selection through an adaptive correctness gate followed by 4D Pareto-based non-dominated sorting.}
\Description{Block diagram of the COEVO loop. A population of Verilog design candidates flows through three stages in sequence: offspring generation, in which LLM-driven evolutionary operators produce new candidates from selected parents; design evaluation, in which each candidate is scored for functional correctness by simulation against an enhanced testbench and for area, delay, and power by logic synthesis; and survivor selection, in which an adaptive correctness gate filters the combined pool before four-dimensional Pareto-based non-dominated sorting reduces it to the next-generation population. The surviving population is fed back as parents for the following generation.}
\label{fig:framework}
\end{figure*}

\section{Problem Formulation}
\label{sec:problem}

Given a natural language hardware specification $\mathcal{S}$, the goal is to generate a Verilog design that achieves both functional correctness and favorable PPA quality. Each design candidate is represented as $I = (V, c, \mathbf{m})$, where $V$ is the Verilog implementation, $c \in [0, 1]$ is the functional correctness score, and $\mathbf{m} = (A, D, P)$ denotes area, critical path delay, and power from logic synthesis.

The ultimate objective is to find a design that is fully functionally correct while minimizing PPA:
\begin{equation}
\label{eq:constrained}
I^{*} = \arg\min_{I} \; \mathbf{m}(I) \quad \text{s.t.} \quad c(I) = 1
\end{equation}
Existing methods treat correctness as a binary prerequisite, optimizing PPA only after $c=1$. COEVO relaxes this into an unconstrained multi-objective optimization:
\begin{equation}
\label{eq:objective}
I^{*} = \arg\min_{I} \big(1 - c(I),\; \mathbf{m}(I) \big)
\end{equation}
where $1 - c$ captures the distance to full correctness and $\mathbf{m}$ is to be minimized component-wise. This relaxation allows the evolutionary process to leverage partially correct but architecturally promising candidates during the search.

To avoid single-scalar collapse, we adopt Pareto dominance to compare candidates, where $I_a$ dominates $I_b$ ($I_a \succ I_b$) if and only if:
\begin{equation}
\label{eq:dominance}
I_a \succ I_b \iff
\begin{cases}
c_a \geq c_b \;\wedge\; \mathbf{m}_a \leq \mathbf{m}_b \\
c_a > c_b \;\lor\; \mathbf{m}_a \neq \mathbf{m}_b
\end{cases}
\end{equation}
where $\mathbf{m}_a \leq \mathbf{m}_b$ denotes component-wise inequality ($A_a \leq A_b \wedge D_a \leq D_b \wedge P_a \leq P_b$). A design is Pareto optimal if no other feasible design dominates it. This formulation places correctness and PPA into a unified objective space, enabling joint optimization rather than separate stages.

\section{COEVO Framework}
\label{sec:framework}

Figure~\ref{fig:framework} illustrates the overall architecture of COEVO. Given a specification $\mathcal{S}$, the framework maintains a population of $N$ candidates and refines them over $G$ generations. Each generation consists of three stages: (1) \textit{offspring generation}, where LLM-driven operators produce $\lambda$ new candidates (Section~\ref{sec:generation}), (2) \textit{design evaluation}, where each candidate is assessed for correctness through simulation and PPA through synthesis (Section~\ref{sec:eval}), and (3) \textit{survivor selection}, where the combined pool is reduced to $N$ individuals through Pareto-based selection (Section~\ref{sec:selection}). After $G$ generations, COEVO returns the Pareto front of the final population.

\subsection{Offspring Generation}
\label{sec:generation}

\subsubsection{Population Initialization}

The initial population $\mathcal{P}_0$ is constructed to ensure broad architectural diversity, providing a wide starting basis for the evolutionary search. Given the specification $\mathcal{S}$, the LLM identifies applicable architecture strategies from a predefined set $\mathcal{K}$ (e.g., behavioral, structural, pipeline, resource-shared, FSM-minimized). For each selected strategy $s_k \in \mathcal{K}$, $\lceil N / |\mathcal{K}| \rceil$ candidates are generated.

\subsubsection{Evolutionary Operators}

In each generation, the offspring set $\mathcal{O}_t$ of $\lambda$ candidates is produced by applying LLM-driven evolutionary operators to parents sampled from the current population $\mathcal{P}_{t-1}$. Each operator constructs a structured prompt containing the specification $\mathcal{S}$, parent information, and operator-specific instructions embedding domain-specific hardware optimization knowledge. COEVO defines seven operators organized into three categories.

\textbf{Correctness-oriented operators} include \textit{Fix} and \textit{Simplify}, both targeting functional improvement. \textit{Fix} takes a single parent along with its diagnostic feedback (failing test cases) and instructs the LLM to correct the identified errors while preserving the overall architecture, such as repairing carry propagation logic without altering pipelining structure. \textit{Simplify} reduces design complexity to escape error-prone implementations, such as replacing deeply nested conditional logic with a cleaner state machine encoding.

\textbf{PPA-oriented operators} include \textit{Optimize}, \textit{Restructure}, and \textit{Explore}, all targeting hardware efficiency. \textit{Optimize} leverages synthesis diagnosis (critical path breakdown, power distribution) to guide targeted improvements, such as applying resource sharing when duplicated arithmetic units are identified. \textit{Restructure} focuses on the critical path, applying transformations such as pipeline insertion or carry-lookahead conversion to reduce delay. \textit{Explore} generates a fundamentally different architecture from the specification alone, promoting population diversity.

\textbf{Joint correctness-PPA operators} include \textit{PPA-aware Fix} and \textit{Architecture Fusion}, both simultaneously addressing functional correctness and PPA. \textit{PPA-aware Fix} repairs functional errors while preserving PPA-beneficial structures identified from synthesis diagnosis, such as pipelined datapaths, restricting modifications to only the logic related to failing test cases. \textit{Architecture Fusion} combines two parents with complementary strengths: one contributing its correctness approach and the other its PPA optimization technique, integrating both advantages into a single offspring.

\subsubsection{Adaptive Operator Selection}

To balance exploration and exploitation, COEVO employs an adaptive selection mechanism based on the Upper Confidence Bound (UCB) algorithm. The selection process consists of two components: score computation and reward assignment. Operators are then sampled proportionally to their scores via softmax. For each operator $a_i$, the selection score is:
\begin{equation}
\label{eq:ucb}
\text{Score}(a_i) = Q(a_i) + c \sqrt{\frac{\ln T}{n_i}}
\end{equation}
where $Q(a_i)$ is the average historical reward, $n_i$ is the selection count of $a_i$, $T$ is the total selections across all operators, and $c$ is the exploration coefficient. The first term favors higher-reward operators (exploitation) while the second encourages less-selected ones (exploration).

The reward signal is category-specific, reflecting each operator category's optimization intent:
\begin{equation}
\label{eq:reward}
r(a_i) = \begin{cases}
1 & \text{if } a_i \in \mathcal{A}_c \text{ and } c' > c_{\text{parent}} \\
1 & \text{if } a_i \in \mathcal{A}_p \text{ and } \mathbf{m}' < \mathbf{m}_{\text{parent}} \wedge c' \geq c_{\text{parent}} \\
1 & \text{if } a_i \in \mathcal{A}_j \text{ and } c' > c_{\text{parent}} \wedge \mathbf{m}' < \mathbf{m}_{\text{parent}} \\
0 & \text{otherwise}
\end{cases}
\end{equation}
where $\mathcal{A}_c$, $\mathcal{A}_p$, and $\mathcal{A}_j$ denote the correctness-oriented, PPA-oriented, and joint operator sets, $c'$ and $\mathbf{m}'$ are the offspring's correctness and PPA, and $c_{\text{parent}}$ and $\mathbf{m}_{\text{parent}}$ are the parent's. Correctness operators are rewarded for any correctness improvement, PPA operators for PPA gains without correctness regression, and joint operators only when both objectives improve. This category-specific design enables the framework to naturally shift resource allocation from correctness-focused to PPA-focused operators as the population matures.

\subsubsection{Synthesis Repair}

Offspring that fail logic synthesis undergo a repair process. The synthesis error diagnosis (e.g., multi-driver conflicts, logic optimized to zero cells) is combined with the specification into a repair prompt. The repair is attempted up to $R$ times, and the repaired candidate replaces the original only if synthesis succeeds without correctness regression.

\subsection{Design Evaluation}
\label{sec:eval}

Each candidate is evaluated along two dimensions: functional correctness and PPA quality.

\subsubsection{Fine-Grained Correctness Evaluation}

A key enabler of co-evolutionary optimization is measuring functional correctness as a continuous quantity rather than a binary outcome. We construct an enhanced testbench for each design that provides both a continuous score and detailed diagnostic feedback. The enhanced testbench is generated by analyzing the specification with the LLM to derive comprehensive test scenarios covering boundary conditions, typical operations, and corner cases. Input stimuli are simulated through a golden reference implementation provided by the benchmark to obtain expected outputs, forming a set of $T$ test cases. The correctness score is defined as:
\begin{equation}
\label{eq:correctness}
c(I) = \frac{p(I)}{T}
\end{equation}
where $p(I)$ is the number of passed test cases. For each failing case, the testbench reports the expected and actual signal values, providing actionable diagnostic feedback for the evolutionary operators and the repair mechanism.

\subsubsection{PPA Evaluation}

Each candidate is submitted to logic synthesis to obtain PPA metrics $\mathbf{m}(I) = (A, D, P)$ representing area, critical path delay, and power. The synthesis log is parsed to extract a structured diagnosis including cell count, critical path composition, and resource utilization, which serves as optimization guidance for PPA-oriented operators. Candidates that fail synthesis are assigned $\mathbf{m} = \bot$. During non-dominated sorting (Section~\ref{sec:selection}), such candidates are treated as dominated on all PPA dimensions, effectively relegating them to lower Pareto levels while still allowing their correctness score to influence selection.

\subsection{Survivor Selection}
\label{sec:selection}

After evaluation, the candidate pool $\mathcal{P}_{t-1} \cup \mathcal{O}_t$ is reduced to $N$ survivors to form the next-generation population $\mathcal{P}_t$. This process consists of two stages: adaptive correctness gating and 4D Pareto-based non-dominated sorting with configurable intra-level ranking.

\subsubsection{Adaptive Correctness Gate}

At each generation $t$, an adaptive threshold $\theta_t$ determines the minimum correctness required for a candidate to enter the selection pool:
\begin{equation}
\label{eq:gate}
\theta_t = \theta_{\min} + (\theta_{\max} - \theta_{\min}) \cdot \left(\frac{t}{G}\right)^{\alpha}
\end{equation}
where $\theta_{\min}$ and $\theta_{\max}$ define the threshold range and $\alpha$ controls the annealing schedule. The gated candidate pool is defined as:
\begin{equation}
\label{eq:gated_pool}
\mathcal{P}^{\theta} = \{I \in \mathcal{P}_{t-1} \cup \mathcal{O}_t \mid c(I) \geq \theta_t\}
\end{equation}

The rationale is twofold. Early generations use a low threshold so that partially correct candidates with favorable PPA can survive and propagate their architectural insights through crossover and mutation. As evolution progresses, the rising threshold steers the population toward full correctness. This smooth transition avoids the abrupt information loss of a hard binary gate, which would immediately discard all partially correct candidates regardless of architectural quality.

\subsubsection{Pareto-Based Non-Dominated Sorting}

The gated pool $\mathcal{P}^{\theta}$ is reduced to $N$ survivors to form $\mathcal{P}_t$ through three steps:

\textbf{(1) Non-dominated sorting.} The pool is partitioned into Pareto levels $\mathcal{F}_1, \mathcal{F}_2, \ldots, \mathcal{F}_L$ via the dominance relation $\succ$ (Eq.~\ref{eq:dominance}), where $\mathcal{F}_1$ is the Pareto front and subsequent levels represent lower quality. This 4D formulation handles the interplay between correctness and PPA: a candidate with $c = 0.95$ and strong PPA may be non-dominated with respect to a candidate with $c = 1.0$ and moderate PPA, preserving both as viable evolutionary building blocks.

\textbf{(2) Intra-level ranking.} Within each Pareto level $\mathcal{F}_k$, individuals are ranked according to a configurable criterion. The default is correctness in descending order, but the criterion can be set to any PPA metric (e.g., power in ascending order for power-oriented optimization) or a secondary non-dominated sorting on a subset of objectives. This configurability allows the framework to prioritize different objectives without modifying its overall structure.

\textbf{(3) Proportional slot allocation.} Survivors are allocated across Pareto levels proportionally to their priority. Each level $\mathcal{F}_k$ receives a number of slots determined by its weight $w_k = 1/(k+1)$:
\begin{equation}
\label{eq:allocation}
s_k = \text{round}\left(\frac{w_k}{\sum_{l=1}^{L} w_l} \cdot N\right)
\end{equation}
Higher-priority levels receive more slots, concentrating selection pressure on the Pareto front while retaining representatives from lower levels for diversity. Slots within each level are filled according to the intra-level ranking. When a level contains fewer candidates than its allocated slots, the surplus cascades to the next level.

\begin{table*}[t]
\centering
\caption{Functional correctness comparison with agentic methods across four LLM backbones. All values are reported as Pass@$k$ (\%). \textbf{Bold} indicates the best result per backbone. {\textcolor{green!70!black}{Green}} numbers indicate improvement over I/O prompting.}
\label{tab:agentic}
\renewcommand{\arraystretch}{0.95}
\setlength{\tabcolsep}{2.8pt}
\footnotesize
\begin{tabular}{l|ccc|ccc||l|ccc|ccc}
\toprule
\multicolumn{7}{c||}{\textbf{GPT-4o-mini}} & \multicolumn{7}{c}{\textbf{GPT-4.1-mini}} \\
\midrule
\multirow{2}{*}{\textbf{Method}} & \multicolumn{3}{c|}{VerilogEval 2.0 (\%)} & \multicolumn{3}{c||}{RTLLM 2.0 (\%)} & \multirow{2}{*}{\textbf{Method}} & \multicolumn{3}{c|}{VerilogEval 2.0 (\%)} & \multicolumn{3}{c}{RTLLM 2.0 (\%)} \\
& Pass@1 & Pass@5 & Pass@10 & Pass@1 & Pass@5 & Pass@10 & & Pass@1 & Pass@5 & Pass@10 & Pass@1 & Pass@5 & Pass@10 \\
\midrule
I/O & 52.4 & 65.2 & 68.6 & 47.2 & 61.3 & 64.0 & I/O & 62.8 & 78.5 & 82.7 & 57.6 & 70.7 & 74.0 \\
VeriOpt & 60.7\up{8.3} & 66.4\up{1.2} & 71.9\up{3.3} & 56.4\up{9.2} & 62.3\up{1.0} & 68.0\up{4.0} & VeriOpt & 74.2\up{11.4} & 81.8\up{3.3} & 84.9\up{2.2} & 67.4\up{9.8} & 74.1\up{3.4} & 76.8\up{2.8} \\
VeriAgent & 72.4\up{20.0} & 77.8\up{12.6} & 80.8\up{12.2} & 67.9\up{20.7} & 73.0\up{11.7} & 76.3\up{12.3} & VeriAgent & 82.1\up{19.3} & 85.3\up{6.8} & 89.8\up{7.1} & 74.3\up{16.7} & 77.5\up{6.8} & 81.9\up{7.9} \\
VerilogCoder & 74.2\up{21.8} & 79.9\up{14.7} & 81.3\up{12.7} & 68.2\up{21.0} & 74.4\up{13.1} & 76.7\up{12.7} & VerilogCoder & 85.8\up{23.0} & 88.1\up{9.6} & 91.6\up{8.9} & 76.1\up{18.5} & 81.2\up{10.5} & 83.3\up{9.3} \\
VeriMoA & 73.4\up{21.0} & 80.5\up{15.3} & 82.6\up{14.0} & 69.6\up{22.4} & 76.3\up{15.0} & 78.2\up{14.2} & VeriMoA & 86.0\up{23.2} & 90.3\up{11.8} & 92.1\up{9.4} & 77.5\up{19.9} & 82.2\up{11.5} & 84.7\up{10.7} \\
REvolution & 77.9\up{25.5} & 82.8\up{17.6} & 83.8\up{15.2} & 74.7\up{27.5} & 78.3\up{17.0} & 80.4\up{16.4} & REvolution & 89.3\up{26.5} & 91.1\up{12.6} & 92.4\up{9.7} & 80.5\up{22.9} & 83.3\up{12.6} & 85.7\up{11.7} \\
EvolVE & 80.7\up{28.3} & 85.6\up{20.4} & 87.5\up{18.9} & 78.1\up{30.9} & 80.8\up{19.5} & 82.6\up{18.6} & EvolVE & 91.3\up{28.5} & 93.3\up{14.8} & 95.2\up{12.5} & 83.8\up{26.2} & 86.2\up{15.5} & 88.7\up{14.7} \\
\textbf{COEVO} & \textbf{86.5}\up{34.1} & \textbf{88.4}\up{23.2} & \textbf{89.2}\up{20.6} & \textbf{83.2}\up{36.0} & \textbf{87.2}\up{25.9} & \textbf{88.6}\up{24.6} & \textbf{COEVO} & \textbf{94.4}\up{31.6} & \textbf{96.8}\up{18.3} & \textbf{97.4}\up{14.7} & \textbf{90.2}\up{32.6} & \textbf{93.7}\up{23.0} & \textbf{94.2}\up{20.2} \\
\midrule
\multicolumn{7}{c||}{\textbf{GPT-5-mini}} & \multicolumn{7}{c}{\textbf{GPT-5.4-mini}} \\
\midrule
\multirow{2}{*}{\textbf{Method}} & \multicolumn{3}{c|}{VerilogEval 2.0 (\%)} & \multicolumn{3}{c||}{RTLLM 2.0 (\%)} & \multirow{2}{*}{\textbf{Method}} & \multicolumn{3}{c|}{VerilogEval 2.0 (\%)} & \multicolumn{3}{c}{RTLLM 2.0 (\%)} \\
& Pass@1 & Pass@5 & Pass@10 & Pass@1 & Pass@5 & Pass@10 & & Pass@1 & Pass@5 & Pass@10 & Pass@1 & Pass@5 & Pass@10 \\
\midrule
I/O & 64.2 & 79.7 & 83.6 & 58.5 & 71.6 & 76.0 & I/O & 66.8 & 81.2 & 85.9 & 64.4 & 80.0 & 82.0 \\
VeriOpt & 78.2\up{14.0} & 82.2\up{2.5} & 86.7\up{3.1} & 71.5\up{13.0} & 74.5\up{2.9} & 79.9\up{3.9} & VeriOpt & 81.2\up{14.4} & 87.9\up{6.7} & 89.7\up{3.8} & 76.3\up{11.9} & 83.6\up{3.6} & 84.2\up{2.2} \\
VeriAgent & 84.9\up{20.7} & 88.6\up{8.9} & 90.7\up{7.1} & 77.1\up{18.6} & 81.3\up{9.7} & 83.9\up{7.9} & VeriAgent & 92.2\up{25.4} & 93.8\up{12.6} & 95.7\up{9.8} & 83.2\up{18.8} & 86.2\up{6.2} & 88.1\up{6.1} \\
VerilogCoder & 89.2\up{25.0} & 92.3\up{12.6} & 94.1\up{10.5} & 78.4\up{19.9} & 82.7\up{11.1} & 84.6\up{8.6} & VerilogCoder & 94.1\up{27.3} & 95.8\up{14.6} & 96.4\up{10.5} & 82.8\up{18.4} & 86.3\up{6.3} & 88.9\up{6.9} \\
VeriMoA & 88.1\up{23.9} & 91.4\up{11.7} & 93.2\up{9.6} & 80.3\up{21.8} & 84.1\up{12.5} & 86.5\up{10.5} & VeriMoA & 93.8\up{27.0} & 95.2\up{14.0} & 96.1\up{10.2} & 84.2\up{19.8} & 88.3\up{8.3} & 89.8\up{7.8} \\
REvolution & 90.4\up{26.2} & 92.5\up{12.8} & 94.9\up{11.3} & 82.4\up{23.9} & 85.9\up{14.3} & 88.2\up{12.2} & REvolution & 93.5\up{26.7} & 94.7\up{13.5} & 96.9\up{11.0} & 85.4\up{21.0} & 88.6\up{8.6} & 90.6\up{8.6} \\
EvolVE & 93.9\up{29.7} & 95.1\up{15.4} & 96.7\up{13.1} & 85.4\up{26.9} & 88.5\up{16.9} & 90.9\up{14.9} & EvolVE & 95.8\up{29.0} & 97.2\up{16.0} & 97.8\up{11.9} & 88.2\up{23.8} & 92.1\up{12.1} & 93.4\up{11.4} \\
\textbf{COEVO} & \textbf{95.1}\up{30.9} & \textbf{97.2}\up{17.5} & \textbf{97.9}\up{14.3} & \textbf{92.1}\up{33.6} & \textbf{94.8}\up{23.2} & \textbf{95.7}\up{19.7} & \textbf{COEVO} & \textbf{97.5}\up{30.7} & \textbf{98.8}\up{17.6} & \textbf{99.3}\up{13.4} & \textbf{94.5}\up{30.1} & \textbf{96.9}\up{16.9} & \textbf{97.6}\up{15.6} \\
\bottomrule
\end{tabular}
\end{table*}

\begin{table*}[!t]
\centering
\footnotesize
\caption{Comparison with training-based methods. All training-based models use 7B-scale open-source LLMs. COEVO uses the same Qwen2.5-Coder-7B backbone without any fine-tuning. \colorbox{yellow!50}{First}, \colorbox{cyan!30}{second}, and \colorbox{orange!40}{third} best results are highlighted.}
\label{tab:sft_rl}
\resizebox{\textwidth}{!}{
\renewcommand{\arraystretch}{0.8}
\setlength{\aboverulesep}{2pt}
\setlength{\belowrulesep}{2pt}
\begin{tabular}{llccccccc}
\toprule
\multicolumn{2}{c}{\textbf{Model}} & \textbf{Size} & \multicolumn{3}{c}{\textbf{VerilogEval 2.0 (\%)}} & \multicolumn{3}{c}{\textbf{RTLLM 2.0 (\%)}} \\
\cmidrule(lr){4-6} \cmidrule(lr){7-9}
& & & \textit{Pass@1} & \textit{Pass@5} & \textit{Pass@10} & \textit{Pass@1} & \textit{Pass@5} & \textit{Pass@10} \\
\midrule
\multirow{4}{*}{SFT}
& RTLCoder-Mistral & 7B & 35.6 & 37.7 & 45.2 & 38.7 & 41.8 & 47.6 \\
& RTLCoder-DeepSeek-Coder & 6.7B & 39.7 & 46.0 & 53.4 & 40.8 & 49.8 & 54.7 \\
& OriGen-DeepSeek-Coder & 7B & 51.2 & 56.8 & 61.7 & 41.1 & 58.5 & 62.7 \\
& HaVen-CodeQwen1.5 & 7B & 55.4 & 62.8 & 68.7 & 51.0 & 60.9 & 64.3 \\
\hline
\multirow{4}{*}{SFT+RL}
& VeriRL-DeepSeek-Coder & 6.7B & 64.6 & 71.8 & 74.3 & 58.6 & 66.3 & 68.8 \\
& VeriRL-CodeQwen2.5 & 7B & \cellcolor{orange!40}66.3 & 73.4 & 75.9 & 61.5 & 68.9 & 71.1 \\
& QiMeng-SALV & 7B & 64.4 & \cellcolor{orange!40}74.9 & \cellcolor{orange!40}77.2 & \cellcolor{orange!40}61.8 & \cellcolor{orange!40}71.4 & \cellcolor{orange!40}76.2 \\
& CodeV-R1 & 7B & \cellcolor{cyan!30}68.6 & \cellcolor{yellow!50}78.5 & \cellcolor{yellow!50}81.4 & \cellcolor{cyan!30}68.4 & \cellcolor{cyan!30}79.2 & \cellcolor{cyan!30}82.1 \\
\hline
Agentic & \textbf{COEVO (Qwen2.5-Coder-7B)} & 7B & \cellcolor{yellow!50}\textbf{72.7} & \cellcolor{cyan!30}\textbf{77.8} & \cellcolor{cyan!30}\textbf{80.2} & \cellcolor{yellow!50}\textbf{74.5} & \cellcolor{yellow!50}\textbf{80.8} & \cellcolor{yellow!50}\textbf{84.2} \\
\bottomrule
\end{tabular}
}
\end{table*}

\begin{table*}[!t]
\centering
\caption{Per-design PPA comparison on RTLLM 2.0 using GPT-5.4-mini. \colorbox{bestgreen}{Green} indicates the best PPA product ($A \times D \times P$) per design. \colorbox{failred}{\ding{55}} indicates no functionally correct and synthesizable design was produced.}
\label{tab:ppa}
\resizebox{\textwidth}{!}{%
\setlength{\tabcolsep}{4pt}
\renewcommand{\arraystretch}{0.6}
\begin{tabular}{l|rrr|rrr|rrr|rrr|rrr}
\toprule
\multirow{2}{*}{\textbf{Design}} & \multicolumn{3}{c|}{\textbf{Reference}} & \multicolumn{3}{c|}{\textbf{EvolVE}} & \multicolumn{3}{c|}{\textbf{VeriAgent}} & \multicolumn{3}{c|}{\textbf{REvolution}} & \multicolumn{3}{c}{\textbf{COEVO}} \\
\cmidrule(lr){2-4} \cmidrule(lr){5-7} \cmidrule(lr){8-10} \cmidrule(lr){11-13} \cmidrule(lr){14-16}
& Area & Delay & Power & Area & Delay & Power & Area & Delay & Power & Area & Delay & Power & Area & Delay & Power \\
\midrule
JC\_counter & \cellcolor{bestgreen}340.48 & \cellcolor{bestgreen}0.10 & \cellcolor{bestgreen}299 & \cellcolor{bestgreen}340.48 & \cellcolor{bestgreen}0.10 & \cellcolor{bestgreen}299 & \cellcolor{bestgreen}340.48 & \cellcolor{bestgreen}0.10 & \cellcolor{bestgreen}299 & \cellcolor{bestgreen}340.48 & \cellcolor{bestgreen}0.10 & \cellcolor{bestgreen}299 & \cellcolor{bestgreen}340.48 & \cellcolor{bestgreen}0.10 & \cellcolor{bestgreen}299 \\
LFSR & \cellcolor{bestgreen}25.00 & \cellcolor{bestgreen}0.14 & \cellcolor{bestgreen}26.7 & 24.47 & 0.16 & 27.2 & 24.47 & 0.16 & 27.2 & 24.47 & 0.16 & 27.2 & \cellcolor{bestgreen}25.00 & \cellcolor{bestgreen}0.14 & \cellcolor{bestgreen}26.7 \\
LIFObuffer & 228.23 & 0.34 & 372 & 228.23 & 0.34 & 372 & 227.70 & 0.36 & 271 & 227.70 & 0.36 & 271 & \cellcolor{bestgreen}197.90 & \cellcolor{bestgreen}0.39 & \cellcolor{bestgreen}226 \\
RAM & 474.81 & 0.22 & 607 & 470.82 & 0.24 & 608 & 447.68 & 0.32 & 454 & 447.68 & 0.32 & 454 & \cellcolor{bestgreen}435.97 & \cellcolor{bestgreen}0.31 & \cellcolor{bestgreen}420 \\
ROM & 21.81 & 0.21 & 10.1 & 21.81 & 0.21 & 10.1 & 21.81 & 0.21 & 10.1 & \cellcolor{bestgreen}20.22 & \cellcolor{bestgreen}0.15 & \cellcolor{bestgreen}14.7 & \cellcolor{bestgreen}20.22 & \cellcolor{bestgreen}0.15 & \cellcolor{bestgreen}14.7 \\
accu & 230.62 & 0.49 & 726 & 230.62 & 0.49 & 726 & 218.39 & 0.49 & 693 & 217.85 & 0.49 & 680 & \cellcolor{bestgreen}231.95 & \cellcolor{bestgreen}0.49 & \cellcolor{bestgreen}152 \\
adder\_16bit & 96.82 & 0.59 & 61.5 & 96.82 & 0.59 & 61.5 & 97.89 & 0.57 & 60.1 & 97.89 & 0.57 & 60.1 & \cellcolor{bestgreen}93.90 & \cellcolor{bestgreen}0.55 & \cellcolor{bestgreen}61.9 \\
adder\_32bit & 208.54 & 0.87 & 138 & 239.40 & 0.64 & 158 & 191.79 & 1.03 & 119 & 239.40 & 0.64 & 158 & \cellcolor{bestgreen}191.79 & \cellcolor{bestgreen}1.02 & \cellcolor{bestgreen}119 \\
adder\_8bit & \cellcolor{bestgreen}48.94 & \cellcolor{bestgreen}0.31 & \cellcolor{bestgreen}29.0 & 55.06 & 0.27 & 34.1 & 55.06 & 0.27 & 34.1 & 55.06 & 0.27 & 34.1 & 55.06 & 0.27 & 34.1 \\
adder\_bcd & 42.83 & 0.31 & 36.5 & 33.52 & 0.29 & 31.8 & 32.45 & 0.28 & 32.9 & 33.52 & 0.29 & 31.8 & \cellcolor{bestgreen}32.19 & \cellcolor{bestgreen}0.28 & \cellcolor{bestgreen}31.0 \\
adder\_pipe\_64bit & 2529.39 & 0.83 & 2910 & 1480.82 & 1.07 & 1630 & 1480.82 & 1.07 & 1630 & 1480.82 & 1.07 & 1630 & \cellcolor{bestgreen}994.31 & \cellcolor{bestgreen}0.66 & \cellcolor{bestgreen}1100 \\
alu & 1952.97 & 1.93 & 847 & 1190.35 & 1.74 & 578 & 1280.79 & 1.35 & 643 & \cellcolor{bestgreen}1191.15 & \cellcolor{bestgreen}1.52 & \cellcolor{bestgreen}566 & 1219.08 & 1.44 & 592 \\
asyn\_fifo & \cellcolor{bestgreen}1099.91 & \cellcolor{bestgreen}0.33 & \cellcolor{bestgreen}1600 & \cellcolor{failred}\ding{55} & \cellcolor{failred}\ding{55} & \cellcolor{failred}\ding{55} & \cellcolor{failred}\ding{55} & \cellcolor{failred}\ding{55} & \cellcolor{failred}\ding{55} & \cellcolor{failred}\ding{55} & \cellcolor{failred}\ding{55} & \cellcolor{failred}\ding{55} & \cellcolor{failred}\ding{55} & \cellcolor{failred}\ding{55} & \cellcolor{failred}\ding{55} \\
barrel\_shifter & \cellcolor{bestgreen}39.37 & \cellcolor{bestgreen}0.17 & \cellcolor{bestgreen}14.1 & 39.10 & 0.17 & 16.0 & 39.10 & 0.17 & 16.0 & 39.10 & 0.17 & 16.0 & 39.10 & 0.17 & 16.0 \\
calendar & 164.12 & 0.42 & 136 & 168.38 & 0.39 & 136 & 152.42 & 0.38 & 136 & 168.38 & 0.39 & 136 & \cellcolor{bestgreen}139.65 & \cellcolor{bestgreen}0.38 & \cellcolor{bestgreen}127 \\
comparator\_3bit & 11.97 & 0.15 & 5.69 & 13.83 & 0.08 & 6.13 & 11.70 & 0.10 & 5.24 & 13.83 & 0.08 & 6.13 & \cellcolor{bestgreen}11.70 & \cellcolor{bestgreen}0.09 & \cellcolor{bestgreen}5.22 \\
comparator\_4bit & 18.89 & 0.16 & 8.91 & 17.02 & 0.12 & 8.19 & 17.29 & 0.12 & 7.60 & 17.02 & 0.12 & 8.19 & \cellcolor{bestgreen}16.49 & \cellcolor{bestgreen}0.11 & \cellcolor{bestgreen}7.09 \\
counter\_12 & 34.85 & 0.24 & 98.6 & 34.85 & 0.24 & 98.6 & 34.85 & 0.24 & 98.6 & 34.85 & 0.24 & 98.6 & \cellcolor{bestgreen}35.91 & \cellcolor{bestgreen}0.16 & \cellcolor{bestgreen}109 \\
div\_16bit & \cellcolor{bestgreen}737.62 & \cellcolor{bestgreen}5.64 & \cellcolor{bestgreen}22600 & 747.46 & 5.33 & 25700 & 748.79 & 5.40 & 24300 & 747.46 & 5.33 & 25700 & \cellcolor{bestgreen}737.62 & \cellcolor{bestgreen}5.64 & \cellcolor{bestgreen}22600 \\
edge\_detect & 18.35 & 0.12 & 18.4 & 18.35 & 0.12 & 18.4 & 18.09 & 0.11 & 18.4 & 18.09 & 0.11 & 18.4 & \cellcolor{bestgreen}12.77 & \cellcolor{bestgreen}0.10 & \cellcolor{bestgreen}10.2 \\
fixed\_point\_adder & 472.95 & 1.23 & 401 & \cellcolor{bestgreen}464.70 & \cellcolor{bestgreen}0.98 & \cellcolor{bestgreen}385 & \cellcolor{bestgreen}464.70 & \cellcolor{bestgreen}0.98 & \cellcolor{bestgreen}385 & 472.95 & 1.23 & 401 & 469.76 & 0.98 & 394 \\
fixed\_point\_subtractor & 587.86 & 1.24 & 487 & 582.27 & 1.13 & 479 & 508.06 & 1.37 & 420 & 582.27 & 1.13 & 479 & \cellcolor{bestgreen}520.03 & \cellcolor{bestgreen}0.99 & \cellcolor{bestgreen}401 \\
float\_multi & 5980.48 & 2.32 & 65900 & 4846.25 & 2.07 & 41300 & 4846.25 & 2.07 & 41300 & 4846.25 & 2.07 & 41300 & \cellcolor{bestgreen}4234.99 & \cellcolor{bestgreen}3.47 & \cellcolor{bestgreen}23900 \\
freq\_div & 116.24 & 0.39 & 110 & 91.24 & 0.32 & 95.9 & 91.24 & 0.30 & 95.8 & 91.24 & 0.32 & 95.9 & \cellcolor{bestgreen}82.19 & \cellcolor{bestgreen}0.34 & \cellcolor{bestgreen}72.3 \\
freq\_divbyeven & 31.65 & 0.21 & 43.7 & 38.57 & 0.27 & 43.9 & \cellcolor{failred}\ding{55} & \cellcolor{failred}\ding{55} & \cellcolor{failred}\ding{55} & \cellcolor{failred}\ding{55} & \cellcolor{failred}\ding{55} & \cellcolor{failred}\ding{55} & \cellcolor{bestgreen}19.15 & \cellcolor{bestgreen}0.17 & \cellcolor{bestgreen}21.7 \\
freq\_divbyfrac & 48.41 & 0.20 & 40.1 & \cellcolor{failred}\ding{55} & \cellcolor{failred}\ding{55} & \cellcolor{failred}\ding{55} & \cellcolor{failred}\ding{55} & \cellcolor{failred}\ding{55} & \cellcolor{failred}\ding{55} & \cellcolor{failred}\ding{55} & \cellcolor{failred}\ding{55} & \cellcolor{failred}\ding{55} & \cellcolor{bestgreen}30.32 & \cellcolor{bestgreen}0.21 & \cellcolor{bestgreen}38.2 \\
freq\_divbyodd & \cellcolor{bestgreen}58.52 & \cellcolor{bestgreen}0.62 & \cellcolor{bestgreen}64.7 & \cellcolor{failred}\ding{55} & \cellcolor{failred}\ding{55} & \cellcolor{failred}\ding{55} & \cellcolor{failred}\ding{55} & \cellcolor{failred}\ding{55} & \cellcolor{failred}\ding{55} & \cellcolor{failred}\ding{55} & \cellcolor{failred}\ding{55} & \cellcolor{failred}\ding{55} & \cellcolor{bestgreen}58.52 & \cellcolor{bestgreen}0.62 & \cellcolor{bestgreen}64.7 \\
fsm & 45.49 & 0.18 & 52.4 & 36.97 & 0.13 & 42.5 & 25.80 & 0.18 & 27.4 & 36.97 & 0.13 & 42.5 & \cellcolor{bestgreen}26.07 & \cellcolor{bestgreen}0.14 & \cellcolor{bestgreen}20.8 \\
instr\_reg & 117.04 & 0.16 & 129 & 117.04 & 0.16 & 129 & 117.04 & 0.16 & 129 & 117.04 & 0.16 & 129 & \cellcolor{bestgreen}117.31 & \cellcolor{bestgreen}0.14 & \cellcolor{bestgreen}129 \\
multi\_16bit & 951.48 & 1.98 & 1250 & 530.14 & 0.79 & 1920 & 530.14 & 0.79 & 1920 & 530.14 & 0.79 & 1920 & \cellcolor{bestgreen}560.73 & \cellcolor{bestgreen}0.74 & \cellcolor{bestgreen}747 \\
multi\_8bit & 525.35 & 1.62 & 852 & 525.35 & 1.62 & 852 & 345.27 & 1.19 & 1360 & 329.04 & 0.74 & 455 & \cellcolor{bestgreen}330.37 & \cellcolor{bestgreen}0.74 & \cellcolor{bestgreen}450 \\
multi\_booth\_8bit & 414.96 & 0.79 & 598 & 414.43 & 0.78 & 600 & 417.89 & 0.75 & 574 & 417.89 & 0.75 & 574 & \cellcolor{bestgreen}265.20 & \cellcolor{bestgreen}0.71 & \cellcolor{bestgreen}186 \\
multi\_pipe\_4bit & 171.57 & 0.34 & 189 & 171.57 & 0.34 & 189 & 171.57 & 0.29 & 196 & 149.23 & 0.31 & 175 & \cellcolor{bestgreen}163.59 & \cellcolor{bestgreen}0.28 & \cellcolor{bestgreen}173 \\
multi\_pipe\_8bit & 866.36 & 0.79 & 854 & 926.48 & 0.47 & 1770 & 866.36 & 0.79 & 854 & 866.36 & 0.79 & 854 & \cellcolor{bestgreen}924.08 & \cellcolor{bestgreen}0.48 & \cellcolor{bestgreen}911 \\
parallel2serial & 48.68 & 0.20 & 46.3 & 48.68 & 0.20 & 46.3 & 47.88 & 0.16 & 56.8 & 44.16 & 0.19 & 41.4 & \cellcolor{bestgreen}19.95 & \cellcolor{bestgreen}0.16 & \cellcolor{bestgreen}25.8 \\
pe & 3649.52 & 1.72 & 35700 & 3649.52 & 1.72 & 35700 & 3649.52 & 1.72 & 35700 & 3651.91 & 1.70 & 35600 & \cellcolor{bestgreen}3392.83 & \cellcolor{bestgreen}1.72 & \cellcolor{bestgreen}30100 \\
pulse\_detect & 29.26 & 0.15 & 34.9 & 29.26 & 0.15 & 34.9 & 28.73 & 0.25 & 15.5 & 11.70 & 0.10 & 9.81 & \cellcolor{bestgreen}6.65 & \cellcolor{bestgreen}0.12 & \cellcolor{bestgreen}5.15 \\
radix2\_div & 414.43 & 0.59 & 362 & \cellcolor{bestgreen}286.75 & \cellcolor{bestgreen}0.35 & \cellcolor{bestgreen}858 & \cellcolor{failred}\ding{55} & \cellcolor{failred}\ding{55} & \cellcolor{failred}\ding{55} & \cellcolor{bestgreen}286.75 & \cellcolor{bestgreen}0.35 & \cellcolor{bestgreen}858 & 414.43 & 0.59 & 362 \\
right\_shifter & \cellcolor{bestgreen}36.18 & \cellcolor{bestgreen}0.08 & \cellcolor{bestgreen}37.5 & \cellcolor{bestgreen}36.18 & \cellcolor{bestgreen}0.08 & \cellcolor{bestgreen}37.5 & \cellcolor{bestgreen}36.18 & \cellcolor{bestgreen}0.08 & \cellcolor{bestgreen}37.5 & \cellcolor{bestgreen}36.18 & \cellcolor{bestgreen}0.08 & \cellcolor{bestgreen}37.5 & \cellcolor{bestgreen}36.18 & \cellcolor{bestgreen}0.08 & \cellcolor{bestgreen}37.5 \\
ring\_counter & 46.82 & 0.10 & 56.2 & 46.82 & 0.10 & 56.2 & 46.82 & 0.10 & 56.2 & 46.82 & 0.10 & 56.2 & \cellcolor{bestgreen}32.45 & \cellcolor{bestgreen}0.18 & \cellcolor{bestgreen}30.1 \\
sequence\_detector & 36.44 & 0.15 & 38.6 & 27.40 & 0.22 & 36.4 & 27.40 & 0.22 & 36.4 & 36.44 & 0.15 & 38.6 & \cellcolor{bestgreen}23.14 & \cellcolor{bestgreen}0.14 & \cellcolor{bestgreen}22.2 \\
serial2parallel & 156.14 & 0.41 & 219 & 163.86 & 0.52 & 169 & 156.14 & 0.41 & 219 & 156.14 & 0.41 & 219 & \cellcolor{bestgreen}154.01 & \cellcolor{bestgreen}0.33 & \cellcolor{bestgreen}157 \\
signal\_generator & 69.16 & 0.37 & 26.5 & 69.96 & 0.33 & 26.6 & 67.30 & 0.34 & 26.5 & 61.18 & 0.31 & 26.4 & \cellcolor{bestgreen}38.84 & \cellcolor{bestgreen}0.22 & \cellcolor{bestgreen}18.5 \\
square\_wave & 102.14 & 0.48 & 252 & 102.41 & 0.48 & 268 & 102.14 & 0.48 & 252 & 97.62 & 0.47 & 257 & \cellcolor{bestgreen}88.84 & \cellcolor{bestgreen}0.45 & \cellcolor{bestgreen}257 \\
sub\_64bit & 406.45 & 2.16 & 272 & 411.77 & 2.09 & 276 & 404.59 & 2.08 & 255 & 404.05 & 2.07 & 254 & \cellcolor{bestgreen}498.48 & \cellcolor{bestgreen}0.87 & \cellcolor{bestgreen}324 \\
synchronizer & 65.97 & 0.17 & 59.9 & 65.97 & 0.17 & 59.9 & 60.65 & 0.17 & 55.2 & 55.33 & 0.17 & 50.4 & \cellcolor{bestgreen}50.01 & \cellcolor{bestgreen}0.14 & \cellcolor{bestgreen}45.2 \\
traffic\_light & 161.20 & 0.37 & 538 & 161.20 & 0.37 & 538 & 161.20 & 0.37 & 538 & 161.20 & 0.37 & 538 & \cellcolor{bestgreen}132.73 & \cellcolor{bestgreen}0.38 & \cellcolor{bestgreen}562 \\
up\_down\_counter & 219.45 & 0.67 & 689 & 219.45 & 0.67 & 689 & 219.45 & 0.67 & 689 & 184.34 & 0.55 & 583 & \cellcolor{bestgreen}160.40 & \cellcolor{bestgreen}0.50 & \cellcolor{bestgreen}372 \\
width\_8to16 & 186.73 & 0.25 & 319 & 186.73 & 0.24 & 319 & 186.73 & 0.25 & 319 & 186.73 & 0.24 & 319 & \cellcolor{bestgreen}180.88 & \cellcolor{bestgreen}0.24 & \cellcolor{bestgreen}311 \\
\bottomrule
\end{tabular}%
}
\end{table*}


\section{Experiments}
\label{sec:experiments}

We conduct comprehensive experiments to evaluate COEVO by addressing five research questions: \textbf{RQ1}: How does COEVO compare with existing agentic and training-based methods in functional correctness? \textbf{RQ2}: How well does COEVO perform in PPA optimization compared to existing methods on synthesizable designs? \textbf{RQ3}: What is the contribution of each core component to the overall performance? \textbf{RQ4}: How does the co-evolutionary process jointly optimize correctness and PPA over generations? \textbf{RQ5}: What is the computational cost of COEVO compared with existing methods?

\subsection{Experimental Setup}
\label{sec:setup}

\textbf{Benchmarks.} We evaluate on two standard spec-to-RTL benchmarks: VerilogEval 2.0 \cite{Pinckney2025RevisitingVerilogEval} (156 design tasks covering combinational and sequential circuits) and RTLLM 2.0 \cite{Lu2024RTLLM} (50 design tasks spanning arithmetic modules, control logic, and datapath circuits). During our experiments, we identified and corrected a small number of specification ambiguities and testbench inconsistencies in both benchmarks to ensure fair evaluation across all methods.

\noindent\textbf{LLM Backbones.} For the agentic method comparison, we evaluate with four OpenAI models: GPT-4o-mini, GPT-4.1-mini, GPT-5-mini, and GPT-5.4-mini. To compare with training-based methods that operate on smaller open-source models, we additionally evaluate COEVO with Qwen2.5-Coder-7B. All baselines are evaluated under the same backbone for fair comparison.

\noindent\textbf{Baselines.} For functional correctness evaluation (RQ1), we compare against two categories of methods. Among agentic methods, we include I/O prompting, VeriOpt \cite{Tasnia2025VeriOpt}, VeriAgent \cite{Wang2026VeriAgent}, VerilogCoder \cite{Ho2025VerilogCoder}, VeriMoA \cite{Ping2025VeriMoA}, REvolution \cite{Min2026REvolution}, and EvolVE \cite{Hsin2026EvolVE}. Among training-based methods, we include SFT models (RTLCoder \cite{Liu2024RTLCoder}, OriGen \cite{Cui2024OriGen}, HaVen \cite{Yang2025HAVEN}) and SFT+RL models (VeriRL \cite{Teng2025VERIRL}, QiMeng-SALV \cite{Zhang2025QiMengSALV}, CodeV-R1 \cite{Zhu2025QiMengCodeVR1}). For PPA comparison (RQ2), we compare against EvolVE, VeriAgent, and REvolution on RTLLM 2.0 using GPT-5.4-mini as the backbone.

\noindent\textbf{Evaluation Metrics.} Functional correctness is measured by Pass@$k$ ($k \in \{1, 5, 10\}$) computed over 10 independent runs per design using the original benchmark testbenches, following the standard unbiased estimator \cite{Ping2025VeriMoA}. PPA quality is evaluated through logic synthesis using Yosys and power analysis using OpenSTA with the NanGate 45nm standard cell library. We report area ($\mu$m$^2$), critical path delay (ns), and power ($\mu$W).

\noindent\textbf{Hyperparameters.} COEVO uses population size $N=10$, offspring count $\lambda=10$, maximum generations $G=10$, and repair budget $R=3$. The adaptive gate parameters are $\theta_{\min}=0.25$, $\theta_{\max}=1.0$, $\alpha=2.0$. The UCB exploration coefficient is $c=2.0$. All LLM calls use temperature 0.8 and top\_p 0.95.

\subsection{Functional Correctness (RQ1)}
\label{sec:rq1}

\textbf{Comparison with Agentic Methods.} Table~\ref{tab:agentic} presents the functional correctness comparison across four LLM backbones on VerilogEval 2.0 and RTLLM 2.0. COEVO achieves the highest Pass@$k$ across all backbone and benchmark configurations. With GPT-5.4-mini, COEVO attains 97.5\% Pass@1 on VerilogEval 2.0, surpassing the strongest baseline EvolVE by 1.7 percentage points. On the more challenging RTLLM 2.0 benchmark, the improvement is particularly pronounced: COEVO consistently outperforms the second-best method by 5.1 to 6.7 points in Pass@1 across all four backbones, indicating that the co-evolutionary mechanism is especially effective on complex designs where correctness and architectural quality are more tightly coupled.

The advantage of COEVO is consistent across LLM backbones. Even with the weakest backbone GPT-4o-mini, COEVO achieves 86.5\% Pass@1 on VerilogEval 2.0, outperforming EvolVE (80.7\%) by 5.8 points and REvolution (77.9\%) by 8.6 points. As the backbone strengthens, COEVO continues to scale effectively, reaching 99.3\% Pass@10 on VerilogEval 2.0 with GPT-5.4-mini.

\noindent\textbf{Comparison with Training-Based Methods.} Table~\ref{tab:sft_rl} compares COEVO against SFT and SFT+RL methods, all operating on 7B-scale open-source models. Without any fine-tuning, COEVO with Qwen2.5-Coder-7B achieves 72.7\% Pass@1 on VerilogEval 2.0 and 74.5\% on RTLLM 2.0, outperforming all training-based baselines including CodeV-R1 (68.6\% and 68.4\%), the strongest SFT+RL method, by 4.1 and 6.1 points respectively. This result demonstrates that COEVO's inference-time evolutionary optimization can surpass the gains from domain-specific model training, while remaining complementary to training-based approaches since fine-tuned models can serve as the backbone LLM within the COEVO framework.

\subsection{PPA Quality (RQ2)}
\label{sec:rq2}

Table~\ref{tab:ppa} presents the per-design PPA comparison on RTLLM 2.0 using GPT-5.4-mini. Among 49 designs (excluding unsynthesizable \texttt{clk\_generator}), COEVO produces functionally correct and synthesizable designs for 48, compared to 46 for EvolVE, 44 for VeriAgent, and 45 for REvolution. On designs such as \texttt{freq\_divbyfrac}, all three baselines fail to produce a correct design, while COEVO succeeds, demonstrating that the co-evolutionary framework maintains functional correctness while pursuing PPA optimization.

We evaluate PPA quality using the composite metric $A \times D \times P$ \cite{Wang2026VeriAgent}, where lower values indicate better overall efficiency. COEVO achieves the best PPA product on 43 out of 49 designs, substantially outperforming the reference implementations (8), REvolution (5), VeriAgent (3), and EvolVE (4). The advantage is particularly evident on designs where targeted optimization yields significant gains: for \texttt{accu}, COEVO reduces power by 79.1\% over the reference while maintaining comparable area and delay, demonstrating the ability to exploit specific optimization opportunities through PPA-oriented operators guided by synthesis diagnosis. For \texttt{fsm}, COEVO achieves 42.7\% area, 22.2\% delay, and 60.3\% power reduction simultaneously, illustrating how the co-evolutionary mechanism enables broad architectural restructuring while preserving correctness through the adaptive gate and joint operators. These improvements stem from the core design of COEVO: allowing PPA-promising candidates with partial correctness to survive and guide the population toward architecturally superior solutions that are subsequently refined to full correctness. The 4D Pareto-based selection further enables COEVO to preserve diverse trade-off solutions across area, delay, and power without collapsing into a single scalar, avoiding the premature convergence caused by weighted-sum fitness aggregation used in existing evolutionary methods.

\subsection{Ablation Study (RQ3)}
\label{sec:rq3}

\begin{figure*}[t]
\centering
\includegraphics[width=0.8\textwidth]{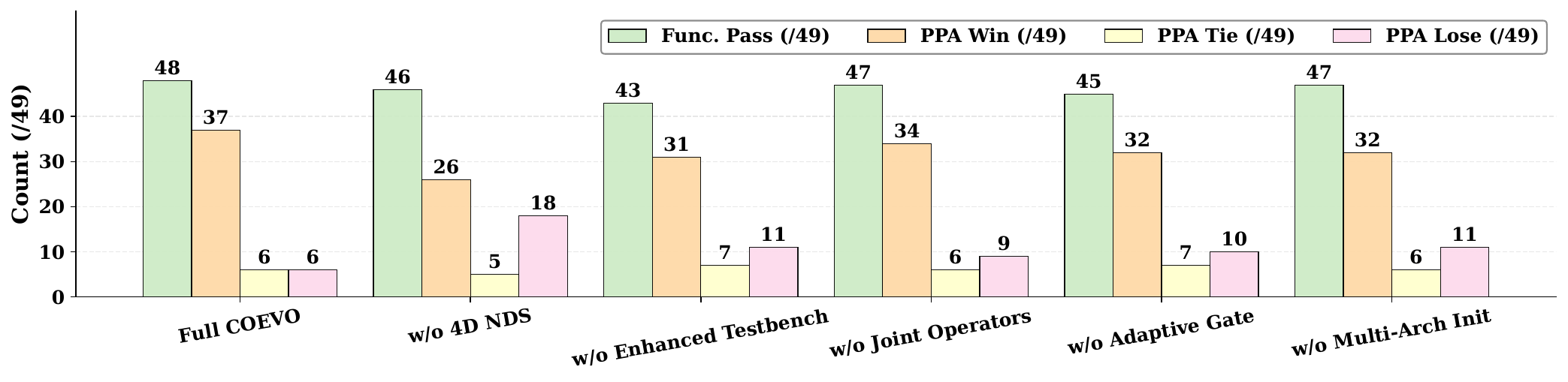}
\caption{Ablation study on RTLLM 2.0 using GPT-5.4-mini. Each group of bars reports Func.\ Pass, PPA Win, PPA Tie, and PPA Lose (out of 49 designs) against the best baseline per design.}
\Description{Grouped bar chart comparing the full COEVO framework against five ablated configurations. For each configuration, four bars report the number of designs out of forty-nine that pass functional verification, and that win, tie, or lose on PPA against the best baseline for that design. The full framework has the highest functional pass count and the highest PPA win count; removing four-dimensional non-dominated sorting produces the largest drop in PPA wins, and removing the enhanced testbench produces the largest drop in functional passes.}
\label{fig:ablation}
\end{figure*}

We evaluate five ablation configurations on RTLLM 2.0 using GPT-5.4-mini, each removing one component from the full COEVO framework: (1) \textbf{w/o 4D NDS}, replacing 4D non-dominated sorting with binary correctness split and weighted-sum PPA fitness, (2) \textbf{w/o Enhanced Testbench}, using the original benchmark testbench with fewer test cases and less detailed feedback, (3) \textbf{w/o Joint Operators}, removing PPA-aware Fix and Architecture Fusion, (4) \textbf{w/o Adaptive Gate}, removing the correctness threshold so all candidates directly enter non-dominated sorting, and (5) \textbf{w/o Multi-Arch Init}, using only a single behavioral strategy for initialization. Figure~\ref{fig:ablation} reports Func.\ Pass and PPA Win/Tie/Lose counts out of 49 designs, where each design is compared against the best result among all baselines and designs that fail functional correctness are counted as PPA Lose.

\textbf{w/o 4D NDS} causes the largest PPA degradation (Win: 37$\to$26, Lose: 6$\to$18), which confirms that 4D Pareto-based non-dominated sorting is the most critical component: collapsing PPA into a single scalar prevents the framework from preserving trade-off diversity across area, delay, and power, leading to suboptimal selection decisions throughout the evolutionary process. \textbf{w/o Enhanced Testbench} produces the largest correctness drop (Func.\ Pass: 48$\to$43, PPA Win: 37$\to$31), as the reduced test case coverage and coarser diagnostic feedback weaken the guidance available to evolutionary operators while diminishing the discriminative power of correctness as a Pareto dimension. \textbf{w/o Adaptive Gate} and \textbf{w/o Multi-Arch Init} both reduce PPA Win to 32, where removing the gate allows low-quality candidates to dilute selection pressure (Func.\ Pass: 48$\to$45) and removing diverse initialization limits the starting architectural coverage (Func.\ Pass: 48$\to$47). \textbf{w/o Joint Operators} shows the smallest degradation (PPA Win: 37$\to$34, Func.\ Pass: 48$\to$47), indicating that the remaining operators can address correctness and PPA, though the absence of cross-objective operators weakens the coupling between the two objectives.

\begin{figure*}[t]
\centering
\includegraphics[width=0.8\textwidth]{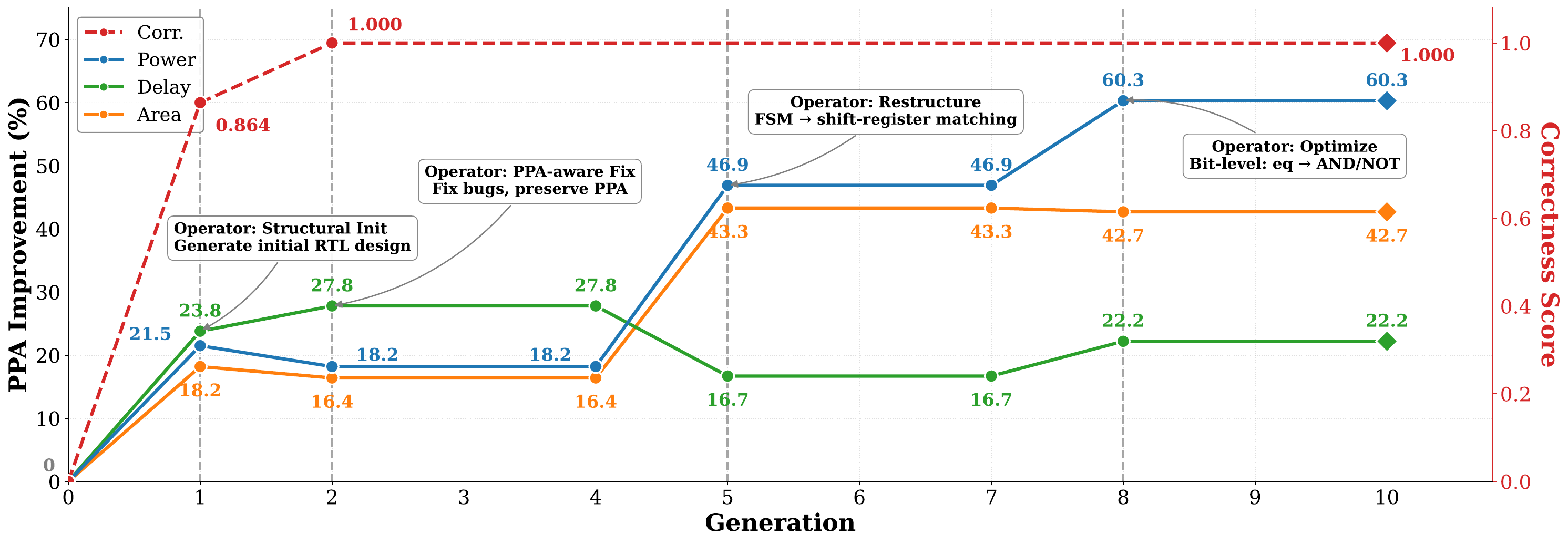}
\caption{Co-evolutionary trajectory of COEVO on \texttt{fsm} from RTLLM 2.0. The left axis shows PPA improvement (\%) over the reference design for area, delay, and power. The right axis shows the correctness score.}
\Description{Line chart tracing ten generations of evolution on the fsm design. Three lines plotted against the left axis show the percentage improvement in area, delay, and power relative to the reference design, all trending upward over the generations. A fourth line plotted against the right axis shows the correctness score rising from roughly 0.86 in the first generation to 1.0 by the second generation and remaining there. Text annotations at four transition points name the evolutionary operator responsible for each change: Structural Init, PPA-aware Fix, Restructure, and Optimize.}
\label{fig:case_study}
\end{figure*}

\subsection{Case Study (RQ4)}
\label{sec:rq4}

Figure~\ref{fig:case_study} traces the co-evolutionary trajectory of a single run on the \texttt{fsm} design from RTLLM 2.0. In generation 1, \textit{Structural Init} produces a five-state FSM with $c=0.864$ and PPA improvements of 18.2\% area, 23.8\% delay, and 21.5\% power. The partially correct candidate survives the adaptive gate and propagates architectural advantages as a stepping stone to later generations. In generation 2, \textit{PPA-aware Fix} repairs the transition logic to $c=1.0$ with minimal structural changes, preserving the inherited PPA quality. In generation 5, \textit{Restructure} replaces the FSM with shift-register-based pattern matching, yielding 43.3\% area and 46.9\% power gains. In generation 8, \textit{Optimize} decomposes the comparator into AND/NOT gate-level logic, pushing power to 60.3\% and delay to 22.2\%. The trajectory shows the mechanisms of COEVO working in concert: the adaptive gate preserves promising candidates, joint operators bridge correctness and PPA, and the search transitions from functional repair to architectural optimization.

\begin{table}[t]
\centering
\caption{Per-design cost and outcome on RTLLM 2.0.}
\label{tab:cost}
\begin{tabular*}{\columnwidth}{@{\extracolsep{\fill}}lrrc@{}}
\toprule
\textbf{Method} & \textbf{Tokens} & \textbf{Pass@1} & \textbf{PPA W/T/L} \\
\midrule
I/O          & 0.9k   & 64.4\% & --     \\
VeriAgent    & 17.0k  & 83.2\% & 0/3/46 \\
VerilogCoder & 110.9k & 82.8\% & --     \\
VeriMoA      & 97.3k  & 84.2\% & --     \\
REvolution   & 159.2k & 85.4\% & 1/4/44 \\
EvolVE       & 174.8k & 88.2\% & 0/4/45 \\
\textbf{COEVO} & 143.1k & \textbf{94.5\%} & \textbf{37/6/6} \\
\bottomrule
\end{tabular*}
\end{table}

\subsection{Cost Analysis (RQ5)}
\label{sec:rq5}

Table~\ref{tab:cost} compares the token cost and the resulting performance of COEVO and representative agentic baselines on RTLLM 2.0 with GPT-5.4-mini. All evolutionary methods share identical hyperparameters ($N{=}\lambda{=}10$, $G{=}10$), the same backbone, and the same synthesis toolchain, while non-evolutionary baselines follow their original configurations. COEVO consumes 143.1k tokens per design, fewer than REvolution (159.2k) and EvolVE (174.8k), while attaining the highest Pass@1 and the largest number of strict PPA wins. In answer to RQ5, COEVO delivers its correctness and PPA gains at a lower token cost than existing evolutionary baselines. The efficiency stems from the co-evolutionary design: partially correct candidates are retained as search signal rather than discarded.

\section{Conclusion}
\label{sec:conclusion}

We presented COEVO, a co-evolutionary framework for LLM-based RTL code generation that jointly optimizes functional correctness and PPA within a unified evolutionary loop. An enhanced testbench provides continuous correctness scoring and lets COEVO treat correctness as a co-optimization dimension alongside area, delay, and power. An annealed adaptive gate keeps partially correct but architecturally promising candidates in the search, and cross-objective operators strengthen the coupling between the two objectives. COEVO further employs 4D Pareto-based non-dominated sorting with configurable intra-level sorting to preserve the trade-off structure without manual weight tuning. Experiments on VerilogEval 2.0 and RTLLM 2.0 demonstrate state-of-the-art functional correctness across four LLM backbones and the best PPA on 43 out of 49 synthesizable RTLLM 2.0 designs. 


\begin{acks}
The authors acknowledge the support by the National Science Foundation (NSF) under the NSF Award 2243104 under the Center for Complex Particle Systems (COMPASS), the CAREER Award CPS-1453860, NSF Mid-Career Advancement Award BCS-2527046, the U.S. Army Research Office (ARO) under Grant No.\ W911NF-23-1-0111, the Defense Advanced Research Projects Agency (DARPA) Young Faculty Award and DARPA Director Fellowship Award under Grant Number N66001-17-1-4044, Intel Faculty Awards, and a Northrop Grumman grant. P.B. is also grateful to the National Institutes of Health (NIH) for Grants R01 AG079957, ``Interpretable machine learning to synergize brain age estimation and neuroimaging genetics,'' and RF1 AG082201, ``Neurovascular calcification and ADRD in two nonindustrial Native American populations.'' The views, opinions, and/or findings in this article are those of the authors and should not be interpreted as official views or policies of the Department of Defense, the National Institutes of Health, or the National Science Foundation.
\end{acks}

\bibliographystyle{ACM-Reference-Format}
\bibliography{sample-base}

@String{Computer = "{IEEE} Computer" }

@preprint{Liu2023ChipNemo,
  author    = {Mingjie Liu and Teodor-Dumitru Ene and Robert Kirby and Chris Cheng and Nathaniel Pinckney and Rongjian Liang and Jonah Alben and Himyanshu Anand and Sanmitra Banerjee and Ismet Bayraktaroglu and Bonita Bhaskaran},
  title     = {ChipNemo: Domain-Adapted LLMs for Chip Design},
  year      = {2023},
  month     = oct,
  archivePrefix = {arXiv},
  eprint    = {2311.00176},
}

@preprint{Chang2023ChipGPT,
  author    = {Kaiyan Chang and Ying Wang and Haimeng Ren and Mengdi Wang and Shengwen Liang and Yinhe Han and Huawei Li and Xiaowei Li},
  title     = {ChipGPT: How Far Are We from Natural Language Hardware Design},
  year      = {2023},
  month     = may,
  archivePrefix = {arXiv},
  eprint    = {2305.14019},
}

@preprint{Yang2025LLMVerilog,
  author    = {Guang Yang and Wei Zheng and Xiang Chen and Dong Liang and Peng Hu and Yukui Yang and Shaohang Peng and Zhi Li and Jian Feng and Xin Wei and Kunyuan Sun},
  title     = {Large Language Model for Verilog Code Generation: Literature Review and the Road Ahead},
  year      = {2025},
  month     = oct,
  archivePrefix = {arXiv},
  eprint    = {2512.00020},
}

@preprint{Xu2025LLMsEDA,
  author    = {Kangwei Xu and Denis Schwachhofer and Jason Blocklove and Ilia Polian and Peter Domanski and Dirk Pfl{\"u}ger and Siddharth Garg and Ramesh Karri and Ozgur Sinanoglu and Johann Knechtel and Zhizi Zhao},
  title     = {Large Language Models ({LLMs}) for Electronic Design Automation ({EDA})},
  year      = {2025},
  month     = aug,
  archivePrefix = {arXiv},
  eprint    = {2508.20030},
}

@Inproceedings{Liu2023VerilogEval,
  author    = {Mingjie Liu and Nathaniel Pinckney and Brucek Khailany and Haoxing Ren},
  title     = {VerilogEval: Evaluating Large Language Models for Verilog Code Generation},
  booktitle = {2023 IEEE/ACM International Conference on Computer Aided Design (ICCAD)},
  year      = {2023},
  month     = oct,
  pages     = {1--8},
  publisher = {IEEE},
}

@Inproceedings{Liu2024OpenLLMRTL,
  author    = {Shang Liu and Yao Lu and Wenji Fang and Mengming Li and Zhiyao Xie},
  title     = {OpenLLM-RTL: Open Dataset and Benchmark for LLM-Aided Design RTL Generation},
  booktitle = {Proceedings of the 43rd IEEE/ACM International Conference on Computer-Aided Design},
  year      = {2024},
  month     = oct,
  pages     = {1--9},
}

@Inproceedings{Lu2024RTLLM,
  author    = {Yao Lu and Shang Liu and Qijun Zhang and Zhiyao Xie},
  title     = {RTLLM: An Open-Source Benchmark for Design RTL Generation with Large Language Model},
  booktitle = {2024 29th Asia and South Pacific Design Automation Conference (ASP-DAC)},
  year      = {2024},
  month     = jan,
  pages     = {722--727},
  publisher = {IEEE},
}

@Article{Pinckney2025RevisitingVerilogEval,
  author    = {Nathaniel Pinckney and Christopher Batten and Mingjie Liu and Haoxing Ren and Brucek Khailany},
  title     = {Revisiting VerilogEval: A Year of Improvements in Large-Language Models for Hardware Code Generation},
  journal   = {ACM Transactions on Design Automation of Electronic Systems},
  volume    = {30},
  number    = {6},
  year      = {2025},
  pages     = {1--20},
}

@Inproceedings{Liu2024RTLCoder,
  author    = {Shang Liu and Wenji Fang and Yao Lu and Qijun Zhang and Hongce Zhang and Zhiyao Xie},
  title     = {RTLCoder: Outperforming GPT-3.5 in Design RTL Generation with Our Open-Source Dataset and Lightweight Solution},
  booktitle = {2024 IEEE LLM Aided Design Workshop (LAD)},
  year      = {2024},
  month     = jun,
  pages     = {1--5},
  publisher = {IEEE},
}

@preprint{Pei2024BetterV,
  author    = {Zehua Pei and Hui-Ling Zhen and Mingxuan Yuan and Yu Huang and Bei Yu},
  title     = {BetterV: Controlled Verilog Generation with Discriminative Guidance},
  year      = {2024},
  month     = feb,
  archivePrefix = {arXiv},
  eprint    = {2402.03375},
}

@Article{Zhao2025CodeV,
  author    = {Yang Zhao and Di Huang and Chongxiao Li and Pengwei Jin and Muxin Song and Yinan Xu and Ziyuan Nan and Mingzhe Gao and Tuo Ma and Lei Qi and Yansong Pan},
  title     = {CodeV: Empowering LLMs with HDL Generation Through Multi-Level Summarization},
  journal   = {IEEE Transactions on Computer-Aided Design of Integrated Circuits and Systems},
  year      = {2025},
  month     = aug,
}

@Inproceedings{Cui2024OriGen,
  author    = {Fan Cui and Chenyang Yin and Kexing Zhou and Youwei Xiao and Guangyu Sun and Qiang Xu and Qipeng Guo and Yun Liang and Xingcheng Zhang and Dawn Song and Dahua Lin},
  title     = {OriGen: Enhancing RTL Code Generation with Code-to-Code Augmentation and Self-Reflection},
  booktitle = {Proceedings of the 43rd IEEE/ACM International Conference on Computer-Aided Design},
  year      = {2024},
  month     = oct,
  pages     = {1--9},
}

@Inproceedings{Deng2025ScaleRTL,
  author    = {Chenhui Deng and Yun-Da Tsai and Guan-Ting Liu and Zhongzhi Yu and Haoxing Ren},
  title     = {ScaleRTL: Scaling LLMs with Reasoning Data and Test-Time Compute for Accurate RTL Code Generation},
  booktitle = {2025 ACM/IEEE 7th Symposium on Machine Learning for CAD (MLCAD)},
  year      = {2025},
  month     = sep,
  pages     = {1--9},
  publisher = {IEEE},
}

@preprint{Zhang2025QiMengSALV,
  author    = {Yang Zhang and Rui Zhang and Jiaming Guo and Lei Huang and Di Huang and Yunpu Zhao and Shuyao Cheng and Pengwei Jin and Chongxiao Li and Zidong Du and Xing Hu and Yunji Chen},
  title     = {QiMeng-SALV: Signal-Aware Learning for Verilog Code Generation},
  year      = {2025},
  month     = oct,
  archivePrefix = {arXiv},
  eprint    = {2510.19296},
}

@preprint{Zhu2025QiMengCodeVR1,
  author    = {Yaoyu Zhu and Di Huang and Hanqi Lyu and Xiaoyun Zhang and Chongxiao Li and Wenxuan Shi and Yutong Wu and Jie Mu and Jiahao Wang and Yang Zhao and Pengwei Jin},
  title     = {QiMeng-CodeV-R1: Reasoning-Enhanced Verilog Generation},
  year      = {2025},
  month     = may,
  archivePrefix = {arXiv},
  eprint    = {2505.24183},
}

@Inproceedings{Teng2025VERIRL,
  author    = {Fu Teng and Miao Pan and Xuhong Zhang and Zhezhi He and Yiyao Yang and Xinyi Chai and Mengnan Qi and Liqiang Lu and Jianwei Yin},
  title     = {{VERIRL}: Boosting the LLM-based Verilog Code Generation via Reinforcement Learning},
  booktitle = {2025 IEEE/ACM International Conference On Computer Aided Design (ICCAD)},
  year      = {2025},
  month     = oct,
  pages     = {1--9},
  publisher = {IEEE},
}

@preprint{Wang2025VeriReason,
  author    = {Yiting Wang and Guoheng Sun and Wanghao Ye and Gang Qu and Ang Li},
  title     = {VeriReason: Reinforcement Learning with Testbench Feedback for Reasoning-Enhanced Verilog Generation},
  year      = {2025},
  month     = may,
  archivePrefix = {arXiv},
  eprint    = {2505.11849},
}

@preprint{Deng2026ACERTL,
  author    = {Chenhui Deng and Zhongzhi Yu and Guan-Ting Liu and Nathaniel Pinckney and Brucek Khailany and Haoxing Ren},
  title     = {{ACE-RTL}: When Agentic Context Evolution Meets RTL-Specialized LLMs},
  year      = {2026},
  month     = feb,
  archivePrefix = {arXiv},
  eprint    = {2602.10218},
}

@preprint{Liu2026VeriSure,
  author    = {Jiale Liu and Taiyu Zhou and Tianqi Jiang},
  title     = {Veri-Sure: A Contract-Aware Multi-Agent Framework with Temporal Tracing and Formal Verification for Correct RTL Code Generation},
  year      = {2026},
  month     = jan,
  archivePrefix = {arXiv},
  eprint    = {2601.19747},
}

@Inproceedings{Zhao2025MAGE,
  author    = {Yujie Zhao and Hejia Zhang and Hanxian Huang and Zhongming Yu and Jishen Zhao},
  title     = {{MAGE}: A Multi-Agent Engine for Automated RTL Code Generation},
  booktitle = {2025 62nd ACM/IEEE Design Automation Conference (DAC)},
  year      = {2025},
  month     = jun,
  pages     = {1--7},
  publisher = {IEEE},
}

@preprint{Ping2025VeriMoA,
  author    = {Heng Ping and Arijit Bhattacharjee and Peiyu Zhang and Shixuan Li and Wei Yang and Anzhe Cheng and Xiaole Zhang and Jesse Thomason and Ali Jannesari and Nesreen Ahmed and Paul Bogdan},
  title     = {{VeriMoA}: A Mixture-of-Agents Framework for Spec-to-HDL Generation},
  year      = {2025},
  month     = oct,
  archivePrefix = {arXiv},
  eprint    = {2510.27617},
}

@preprint{Ping2026POET,
  author    = {Heng Ping and Peiyu Zhang and Zhenkun Wang and Shixuan Li and Anzhe Cheng and Wei Yang and Paul Bogdan and Shahin Nazarian},
  title     = {{POET}: Power-Oriented Evolutionary Tuning for LLM-Based RTL PPA Optimization},
  year      = {2026},
  month     = mar,
  archivePrefix = {arXiv},
  eprint    = {2603.19333},
}

@Inproceedings{Tasnia2025VeriOpt,
  author    = {Kimia Tasnia and Alexander Garcia and Tasnuva Farheen and Sazadur Rahman},
  title     = {VeriOpt: PPA-Aware High-Quality Verilog Generation via Multi-Role LLMs},
  booktitle = {2025 IEEE/ACM International Conference On Computer Aided Design (ICCAD)},
  year      = {2025},
  month     = oct,
  pages     = {1--9},
  publisher = {IEEE},
}

@preprint{Chen2025ChipSeekR1,
  author    = {Zhirong Chen and Kaiyan Chang and Zhuolin Li and Xinyang He and Chujie Chen and Cangyuan Li and Mengdi Wang and Haobo Xu and Yinhe Han and Ying Wang},
  title     = {ChipSeek-R1: Generating Human-Surpassing RTL with LLM via Hierarchical Reward-Driven Reinforcement Learning},
  year      = {2025},
  month     = jul,
  archivePrefix = {arXiv},
  eprint    = {2507.04736},
}

@Inproceedings{Gubbi2025PromptingPower,
  author    = {Kevin Immanuel Gubbi and Marcus Halm and Sarbani Kumar and Arvind Sudarshan and Pavan Dheeraj Kota and Mohammadnavid Tarighat and Avesta Sasan and Houman Homayoun},
  title     = {Prompting for Power: Benchmarking Large Language Models for Low-Power RTL Design Generation},
  booktitle = {2025 ACM/IEEE 7th Symposium on Machine Learning for CAD (MLCAD)},
  year      = {2025},
  month     = sep,
  pages     = {1--7},
  publisher = {IEEE},
}

@preprint{Wang2025SymRTLO,
  author    = {Yiting Wang and Wanghao Ye and Ping Guo and Yexiao He and Ziyao Wang and Bowei Tian and Shwai He and Guoheng Sun and Ziqian Shen and Shaohan Chen and Anurag Srivastava and Ang Li},
  title     = {{SymRTLO}: Enhancing RTL Code Optimization with LLMs and Neuron-Inspired Symbolic Reasoning},
  year      = {2025},
  month     = apr,
  archivePrefix = {arXiv},
  eprint    = {2504.10369},
}

@Inproceedings{Thorat2025LLMVeriPPA,
  author    = {Kiran Thorat and Jiahui Zhao and Yaotian Liu and Amit Hasan and Hongwu Peng and Xi Xie and Bin Lei and Caiwen Ding},
  title     = {{LLM-VeriPPA}: Power, Performance, and Area Optimization Aware Verilog Code Generation with Large Language Models},
  booktitle = {2025 ACM/IEEE 7th Symposium on Machine Learning for CAD (MLCAD)},
  year      = {2025},
  month     = sep,
  pages     = {1--7},
  publisher = {IEEE},
}

@preprint{Wang2026VeriAgent,
  author    = {Yaoxiang Wang and Qi Shi and ShangZhan Li and Qingguo Hu and Xinyu Yin and Bo Guo and Xu Han and Maosong Sun and Jinsong Su},
  title     = {{VeriAgent}: A Tool-Integrated Multi-Agent System with Evolving Memory for PPA-Aware RTL Code Generation},
  year      = {2026},
  month     = mar,
  archivePrefix = {arXiv},
  eprint    = {2603.17613},
}

@Inproceedings{Wei2026VFlow,
  author    = {Yangbo Wei and Zhen Huang and Lei He and Li Huang and Ting-Jung Lin and Wei W. Xing},
  title     = {{VFlow}: Discovering Optimal Agentic Workflows for Verilog Generation},
  booktitle = {2026 31st Asia and South Pacific Design Automation Conference (ASP-DAC)},
  year      = {2026},
  month     = jan,
  pages     = {355--361},
  publisher = {IEEE},
}

@Inproceedings{Min2026REvolution,
  author    = {Kyungjun Min and Kyumin Cho and Junhwan Jang and Seokhyeong Kang},
  title     = {{REvolution}: An Evolutionary Framework for RTL Generation Driven by Large Language Models},
  booktitle = {2026 31st Asia and South Pacific Design Automation Conference (ASP-DAC)},
  year      = {2026},
  month     = jan,
  pages     = {282--288},
  publisher = {IEEE},
}

@preprint{Hsin2026EvolVE,
  author    = {Wei-Po Hsin and Ren-Hao Deng and Yao-Ting Hsieh and En-Ming Huang and Shih-Hao Hung},
  title     = {{EvolVE}: Evolutionary Search for LLM-based Verilog Generation and Optimization},
  year      = {2026},
  month     = jan,
  archivePrefix = {arXiv},
  eprint    = {2601.18067},
}

@preprint{Qi2024EvoT,
  author    = {Biqing Qi and Zhouyi Qian and Yiang Luo and Junqi Gao and Dong Li and Kaiyan Zhang and Bowen Zhou},
  title     = {Evolution of Thought: Diverse and High-Quality Reasoning via Multi-Objective Optimization},
  year      = {2024},
  month     = nov,
  archivePrefix = {arXiv},
  eprint    = {2412.07779},
}

@Article{Deb2002NSGAII,
  author    = {Kalyanmoy Deb and Amrit Pratap and Sameer Agarwal and T. A. M. T. Meyarivan},
  title     = {A Fast and Elitist Multiobjective Genetic Algorithm: {NSGA-II}},
  journal   = {IEEE Transactions on Evolutionary Computation},
  volume    = {6},
  number    = {2},
  year      = {2002},
  pages     = {182--197},
}

@Inproceedings{Ho2025VerilogCoder,
  author    = {Chia-Tung Ho and Haoxing Ren and Brucek Khailany},
  title     = {VerilogCoder: Autonomous Verilog Coding Agents with Graph-Based Planning and Abstract Syntax Tree ({AST})-Based Waveform Tracing Tool},
  booktitle = {Proceedings of the AAAI Conference on Artificial Intelligence},
  volume    = {39},
  year      = {2025},
  month     = apr,
  pages     = {300--307},
}

@Inproceedings{Yao2024RTLRewriter,
  author    = {Xufeng Yao and Yiwen Wang and Xing Li and Yingzhao Lian and Ran Chen and Lei Chen and Mingxuan Yuan and Hong Xu and Bei Yu},
  title     = {{RTLRewriter}: Methodologies for Large Models Aided RTL Code Optimization},
  booktitle = {Proceedings of the 43rd IEEE/ACM International Conference on Computer-Aided Design},
  year      = {2024},
  month     = oct,
  pages     = {1--7},
}

@Article{RomeraParedes2024FunSearch,
  author    = {Bernardino Romera-Paredes and Mohammadamin Barekatain and Alexander Novikov and Matej Balog and M. Pawan Kumar and Emilien Dupont and Francisco J. R. Ruiz and Jordan S. Ellenberg and Pengming Wang and Omar Fawzi and Pushmeet Kohli and Alhussein Fawzi},
  title     = {Mathematical Discoveries from Program Search with Large Language Models},
  journal   = {Nature},
  volume    = {625},
  number    = {7995},
  year      = {2024},
  pages     = {468--475},
}

@preprint{Liu2024EoH,
  author    = {Fei Liu and Xialiang Tong and Mingxuan Yuan and Xi Lin and Fu Luo and Zhenkun Wang and Zhichao Lu and Qingfu Zhang},
  title     = {Evolution of Heuristics: Towards Efficient Automatic Algorithm Design Using Large Language Model},
  year      = {2024},
  month     = jan,
  archivePrefix = {arXiv},
  eprint    = {2401.02051},
}

@Inproceedings{Yang2025HAVEN,
  author    = {Yiyao Yang and Fu Teng and Pengju Liu and Mengnan Qi and Chenyang Lv and Ji Li and Xuhong Zhang and Zhezhi He},
  title     = {{HAVEN}: Hallucination-Mitigated LLM for Verilog Code Generation Aligned with HDL Engineers},
  booktitle = {2025 Design, Automation \& Test in Europe Conference (DATE)},
  year      = {2025},
  month     = mar,
  pages     = {1--7},
  publisher = {IEEE},
}

@preprint{yang2025toward,
  author = {Wei Yang and Muyan Weng and Jiacheng Pang and Defu Cao and Heng Ping and Peiyu Zhang and Shixuan Li and Yue Zhao and Qiang Yang and Mengdi Wang and Yan Liu},
  title  = {Toward Evolutionary Intelligence: LLM-Based Agentic Systems with Multi-Agent Reinforcement Learning},
  note   = {Available at SSRN 5819182},
  year   = {2025},
}

@Inproceedings{Ping2025HDLCoRe,
  author    = {Heng Ping and Shixuan Li and Peiyu Zhang and Anzhe Cheng and Shukai Duan and Nikos Kanakaris and Xiongye Xiao and Wei Yang and Shahin Nazarian and Andrei Irimia and Paul Bogdan},
  title     = {{HDLCoRe}: A Training-Free Framework for Mitigating Hallucinations in LLM-Generated HDL},
  booktitle = {2025 IEEE International Conference on LLM-Aided Design (ICLAD)},
  year      = {2025},
  month     = jun,
  pages     = {108--116},
  publisher = {IEEE},
}

\end{document}